\documentclass[sigconf]{acmart}
\AtBeginDocument{%
  }
\copyrightyear{2023} 
\acmYear{2023} 
\setcopyright{acmlicensed}\acmConference[MM '23]{Proceedings of the 31st ACM International Conference on Multimedia}{October 29-November 3, 2023}{Ottawa, ON, Canada}
\acmBooktitle{Proceedings of the 31st ACM International Conference on Multimedia (MM '23), October 29-November 3, 2023, Ottawa, ON, Canada}
\acmPrice{15.00}
\acmDOI{10.1145/3581783.3611877}
\acmISBN{979-8-4007-0108-5/23/10}

\acmSubmissionID{837}

\usepackage{CJKutf8}

\usepackage[utf8]{inputenc} 
\usepackage{url}            
\usepackage{nicefrac}       
\usepackage{lipsum}		
\usepackage{doi}

\usepackage{xcolor}         
\usepackage{microtype} %
\usepackage{balance}

\usepackage{courier}

\usepackage{makecell}
\usepackage{graphicx}
\usepackage{color}
\usepackage{amsfonts}
\usepackage{amsmath}

\usepackage{multirow}
\usepackage{booktabs}
\usepackage{wrapfig}

\def\ie{\mbox{\textit{i.e.}, }}
\def\eg{\mbox{\textit{e.g.}, }}

\def\mD{{\mathcal D}}

\def\mL{{\mathcal L}}

\def\mP{{\mathcal P}}

\DeclareMathAlphabet\mathbfcal{OMS}{cmsy}{b}{n}

\def\0{{\bf 0}}
\usepackage{ dsfont }
\def\1{\mathds{1}}

\def\bs{{\bf s}}

\def\by{{\bf y}}

\def\mmE{{\textbf E}}

\def\exp{{\mathrm{exp}}}
\def\simm{{\mathrm{sim}}}

\def\by{{\bf y}}

\newtheorem*{*thm}{Theorem}

\newtheorem*{*lemma}{Lemma}

\usepackage{enumitem}

\usepackage{array}
\usepackage{tabu}
\makeatletter
\newcommand{\nipstophline}{%
	\noalign {\ifnum 0=`}\fi \hrule height 4pt
	\futurelet \reserved@a \@xhline
}
\newcommand{\nipsbottomhline}{%
	\noalign {\ifnum 0=`}\fi \hrule height 1pt
	\futurelet \reserved@a \@xhline
}

\def\ljc{\textcolor{black}}
\def\jc{\textcolor{black}}

\def\xie{\textcolor{black}}

\def\blue{\textcolor{black}}
\def\ldw{\textcolor{black}}
\def\mmxie{\textcolor{black}}
\def\mmljc{\textcolor{black}}
\def\mmcr{\textcolor{black}}

\def\databone{benchmark}
\def\pretraindata{Zero}
\def\pretrainmodel{R2D2}

\begin{document}

\title{CCMB: A Large-scale Chinese Cross-modal Benchmark}

\renewcommand{\shorttitle}{CCMB: A Large-scale Chinese Cross-modal Benchmark}

\author{Chunyu Xie}
\affiliation{%
  \institution{360 AI Research}
  \city{Beijing}
  \country{China}
}
\email{xiechunyu@360.cn}

\author{Heng Cai}
\authornote{Both are second authors}
\affiliation{
  \institution{360 AI Research}
  \city{Beijing}
  \country{China}
}
\email{caiheng1@360.cn}

\author{Jincheng Li}
\authornotemark[1]
\affiliation{
  \institution{360 AI Research}
  \city{Beijing}
  \country{China}
}
\email{lijincheng@360.cn}

\author{Fanjing Kong}
\affiliation{
  \institution{360 AI Research}
  \city{Beijing}
  \country{China}
}
\email{kongfanjing@360.cn}

\author{Xiaoyu Wu}
\affiliation{
  \institution{360 AI Research}
  \city{Beijing}
  \country{China}
}
\email{wuxiaoyu1@360.cn}

\author{Jianfei Song}
\affiliation{
  \institution{360 AI Research}
  \city{Beijing}
  \country{China}
}
\email{songjianfei@360.cn}

\author{Henrique Morimitsu}
\affiliation{
  \institution{360 AI Research}
  \institution{Tsinghua University}
  \city{Beijing}
  \country{China}
}
\email{henrique.morimitsu@mail.tsinghua.edu.cn}

\author{Lin Yao}
\affiliation{
  \institution{360 AI Research}
  \city{Beijing}
  \country{China}
}
\email{yaolin@360.cn}

\author{Dexin Wang}
\affiliation{
  \institution{360 Search Department}
  \city{Beijing}
  \country{China}
}
\email{wangdexin@360.cn}

\author{Xiangzheng Zhang}
\affiliation{
  \institution{360 Search Department}
  \city{Beijing}
  \country{China}
}
\email{zhangxiangzheng@360.cn}

\author{Dawei Leng}
\affiliation{
  \institution{360 AI Research}
  \city{Beijing}
  \country{China}
}
\email{lengdawei@360.cn}

\author{Baochang Zhang}
\affiliation{
  \institution{Beihang University}
  \city{Beijing}
  \country{China}
}
\email{bczhang@buaa.edu.cn}

\author{Xiangyang Ji}
\affiliation{
  \institution{Tsinghua University}
  \city{Beijing}
  \country{China}
}
\email{xyji@tsinghua.edu.cn}

\author{Yafeng Deng}
\authornote{Corresponding author}
\affiliation{
  \institution{360 AI Research}
  \institution{Tsinghua University}
  \city{Beijing}
  \country{China}
}
\email{dengyafeng@gmail.com}

\renewcommand{\shortauthors}{Chunyu Xie et al.}

\begin{abstract}
    Vision-language pre-training (VLP) on large-scale datasets has shown premier performance on various downstream tasks. In contrast to plenty of available benchmarks with English corpus, large-scale pre-training datasets and downstream datasets with Chinese corpus remain largely unexplored. In this work, we build a large-scale high-quality Chinese Cross-Modal Benchmark named CCMB for the research community, which contains the currently largest public pre-training dataset Zero and five human-annotated fine-tuning datasets for downstream tasks. Zero contains 250 million images paired with 750 million text descriptions, plus two of the five fine-tuning datasets are also currently the largest ones for Chinese cross-modal downstream tasks. Along with the CCMB, we also develop a VLP framework named R2D2, applying a pre-\textbf{R}anking + \textbf{R}anking strategy to learn powerful vision-language representations and a two-way distillation method (i.e., target-guided \textbf{D}istillation and feature-guided \textbf{D}istillation) to further enhance the learning capability. With the Zero and the R2D2 VLP framework, we achieve state-of-the-art performance on twelve downstream datasets from five broad categories of tasks including image-text retrieval, image-text matching, image caption, text-to-image generation, and zero-shot image classification. The datasets, models, and codes are available at \url{https://github.com/yuxie11/R2D2}
    
\end{abstract}



\maketitle

\section{Introduction}

Vision-language pre-training (VLP) mainly learns the semantic correspondence between vision and natural language.
Previous works~\cite{visualbert,FLAVA,VL-T5,OFA} explore the VLP model and achieve significant improvement on various vision-language (V+L) tasks. These methods are supported by massive data~\cite{LAION-400M}, excellent architectures such as Transformer~\cite{Transformer}, and cross-modal models such as CLIP~\cite{CLIP}.

There are plenty of available benchmarks with English corpus, such as Conceptual
Captions \cite{CC}, SBU Captions \cite{SBU}, and LAION \cite{LAION-400M}. Differently, large-scale pre-training datasets and downstream datasets with Chinese corpus are relatively few.
M6-Corpus~\cite{M6} is a multi-modal pre-training dataset in Chinese but not publicly available. BriVL \mmcr{(also called WenLan)}~\cite{wenlan2} constructs a vision-language dataset called WSCD, but only releases 5M image-text pairs. Wukong~\cite{gu2022wukong} is a newly published pre-training dataset with 100M image-text pairs.
Most existing downstream Chinese datasets mainly focus on retrieval tasks, such as Flickr30k-CN~\cite{flickr30k-cn} and COCO-CN~\cite{coco-cn}, which are not sufficient for a complete evaluation of VLP models.
Besides, Flickr30k-CN tries to translate English cross-modal downstream datasets into Chinese, \mmljc{which}, however, fails to cover Chinese idioms and often causes translation errors.

In this paper, we introduce a large-scale Chinese cross-modal benchmark \blue{called CCMB}, including a pre-training dataset (\mmcr{Zero}) and five downstream datasets.
\blue{Specifically, \mmcr{Zero} consists of 250 million images and 750 million descriptive texts, which is the largest public Chinese V+L pre-training dataset. \mmcr{Zero} is collected from the search engine with images and corresponding textual descriptions, by filtering from 5 billion image-text data by user click-through rate (CTR).}
Compared to existing pre-training datasets, \mmcr{Zero} is \ldw{high-quality due to the user CTR filtering method and the diverse textual information for each image.} \blue{Table~\ref{pretrain_dataset} shows an overview of V+L pre-training datasets.}
Together with the pre-training dataset, we provide 5 high-quality \blue{human-annotated} downstream datasets.
\mmxie{To the best of our knowledge, two of them are the first proposed datasets for the Chinese image-text matching task, which is also important for evaluating VLP models. They are also the largest Chinese V+L downstream datasets.}
For the image-text retrieval task, we provide 3 datasets, 
especially our Flickr30k-CNA, which is a more comprehensive and accurate human-annotated dataset than Flickr30k-CN~\cite{flickr30k-cn}. \blue{The statistics of the public and our proposed downstream datasets are shown in Table~\ref{downnstream_dataset}.}

\mmxie{From the perspective of cross-modal learning, existing methods are mainly categorized as single-stream and dual-stream.} Most single-stream methods (\eg \cite{uniter,UNIMO,imagebert}) employ an extra object detector to extract the patch embedding and then align patches and words. As illustrated in \cite{ALBEF}, object detectors are annotation-expensive and computing-expensive, because they require bounding box annotations during pre-training and high-resolution (such as $600 \times 1000$) images during inference.
On the other hand, for dual-stream architectures (\eg \cite{CLIP,ALIGN,wenlan2}), it is non-trivial to model the fine-grained associations between image and text, since the corresponding representations reside in their own semantic space. 
Some works \cite{BLIP,ALBEF,FLAVA} omit the object detection module and combine dual-stream architecture with single-stream architecture, showing powerful performance on multi-modal downstream tasks.

Inspired by this line of work, we introduce a VLP framework \mmcr{called R2D2}, a combination architecture of dual-stream and single-stream.
\mmcr{We apply global contrastive pre-ranking to obtain image-text representations and fine-grained ranking to further improve model performance.}
Besides, we introduce a two-way distillation method into the model, consisting of target-guided distillation and feature-guided distillation. The target-guided distillation increases the robustness when learning from noisy labels, while feature-guided distillation aims to improve the generalization performance.
We apply \mmcr{masked language modeling with enhanced training}, which improves the capability of the model while reducing the training cost.
To summarize, our main contributions are as follows:


%
\begin{itemize}[leftmargin=*]
    \item We construct the largest public Chinese vision-language pre-training dataset, containing 250 million images and 750 million corresponding texts. It is high-quality due to the filtering method by user CTR and the diverse textual information for each image.
    We provide five human-annotated cross-modal downstream datasets, two of which are currently the largest Chinese vision-language downstream datasets.
    \item \mmxie{We introduce a vision-language pre-training framework named \mmcr{R2D2} for cross-modal learning. Our proposed method achieves state-of-the-art performance on \blue{twelve downstream datasets from five broad categories of vision-language tasks,} showing the superior ability of our pre-trained model.}
\end{itemize}


\section{Related Work}

\subsection{Vision-Language Datasets}

Chinese vision-language \databone{} requires images and high-quality Chinese texts, which are hard to obtain and still rare for the research community's reach.
To this end, existing public datasets~\cite{flickr30k-cn, coco-cn} use machine translation to adapt their English versions~\cite{cococaption,Flickr} to Chinese, but the data quality is sacrificed due to machine translation errors.
Newly reported datasets with Chinese texts~\cite{wenlan2,gu2022wukong,M6} are proposed for Chinese VLP. However, they are either not publicly available or lack sufficient downstream tasks. In this paper, we propose a Chinese vision-language \databone{} that covers a large-scale pre-training dataset and five downstream datasets.

\subsection{Vision-Language Pre-training Learning}

The vision-language pre-training architectures can be categorized as: single-stream and dual-stream.
Most existing single-stream models~\cite{uniter,unicodervl,visualbert,12in1,imagebert} concatenate image and text as a single input to model the interactions between image and text within a transformer model~\cite{Transformer}.
On the other hand, popular dual-stream models~\cite{vse,scaling,visualitm,ViLBERT,CLIP,ALIGN,chinese-clip} aim to align image and text into a unified semantic space via contrastive learning.
Besides, some works~\cite{BLIP,ALBEF,FLAVA} align the individual features of images and texts in a dual-stream architecture, and then fuse the features in a unified semantic space via a single-stream architecture.
These works show that a combined architecture of dual-stream and single-stream achieves better performance than only \mmcr{one. R2D2 explores the effective signals via an image-text cross encoder and a text-image cross encoder while also maintaining the bottom dual-stream architecture. Moreover, we improve masked language modeling with enhanced training and 
propose a two-way distillation to stabilize the model representations for vision-language pre-training.}

\section{CCMB}
\label{Collected_Dataset}

\subsection{Pre-training Dataset}\label{sec pretraindata}
Existing public pre-training datasets suffer from two limitations.
First, the image-text pairs are collected usually by their co-occurrence relationship coarsely from third-party search engines or websites. Thus, the collected pairs are inherently noisy.
Second, the text corpus lacks diversity as each image usually has one corresponding text description. 
To overcome these drawbacks, we collect a new dataset for Chinese image-text pre-training, called \pretraindata{}.

\mmljc{To this end, we first collect 5 billion image-text data from an image search engine. We try to mitigate the noise of these image-text pairs via user click-through rate (CTR) and obtain about 250 million images and 750 million corresponding texts, namely \pretraindata{}. In other words, each image in \pretraindata{} is with about 3 textual descriptions, \ie ``Title'', ``Content'', and ``ImageQuery''.}
\mmxie{``Title'' and ``Content'' come from the source webpage containing the image. ``Title'' is the title of the webpage, and ``Content'' represents the surrounding text of the image in the webpage.
``ImageQuery'' is the user search query for the corresponding image.}
The average length of ``Title'', ``Content'', and ``ImageQuery'' is 18, 29, and 5, respectively.
We show an example in Figure~\ref{fig:pretraindata_one_image} and more examples in Appendix.

\begin{table}[t]
\centering 
\caption{\blue{Statistics of the vision-language pre-training datasets. The details of \mmcr{Zero} can refer to Section~\ref{sec pretraindata}.}}
\resizebox{0.48\textwidth}{!}{
\begin{tabular}{lcccc} 
\toprule
Dateset & Language & Availability & \#Image & \#Text
\\
\midrule
Visual Genome~\cite{VG} & English & Yes & 108K & 5.4M \\
SBU Captions~\cite{SBU} & English & Yes & 875K & 875K \\
CC3M~\cite{CC} & English & Yes & 3.1M & 3.1M \\
CC12M~\cite{CC12M} & English & Yes & 12M & 12M \\
RedCaps~\cite{redcaps} & English & Yes & 12M & 12M \\
WIT~\cite{WIT} & Multilingual & Yes & 11.5M & 37.6M \\
YFCC100M~\cite{YFCC100M} & English & Yes & 100M & 200M \\
LAION-400M~\cite{LAION-400M} & English & Yes & 400M & 400M \\
\midrule
WSCD~\cite{wenlan2} & Chinese & Yes & 5M & 5M \\
M6-Corpus~\cite{M6} & Chinese & No & 60.5M  & 60.5M \\
Wukong~\cite{gu2022wukong} & Chinese & Yes & 100M & 100M \\
\mmcr{Zero} & Chinese & Yes & 250M & 750M \\

\bottomrule 
\end{tabular}
} 
\label{pretrain_dataset}
\end{table}

\mmljc{\textbf{How to remove the irrelevant content?}}
\mmxie{
We apply a series of filtering strategies to construct the \mmcr{Zero}. For images, we filter out images with dimensions smaller than 100 pixels or aspect ratio out of the range [1/4, 4]. We then filter images that contain sensitive information, such as sexual, violent scenes, etc. For texts, 
we remove texts shorter than 2 words or longer than 128 words.
Moreover, we remove image-text pairs that contain sensitive words and personal names in the text.}

\mmljc{
\textbf{Why use CTR instead of random selection?}
After collecting 5 billion image-text data, we can randomly select a part of them to conduct pre-training experiments under the consideration of computational resources and time cost. However, random selection brings noises in pre-training, which may degrade model performance. To address this, we use an inherent metric CTR in search engine data. The CTR indicates the number of times that users click on an image for a given text.
We observe that image-text pairs with low CTR are irrelevant in most cases. That is, an image-text pair with high CTR is strongly correlated due to user interaction.
We then rank all 5 billion image-text data by CTR and filter the top 250 million images and the corresponding texts as \pretraindata{}.
}

\mmljc{\textbf{Why use three types of text instead of one?}}
\mmljc{Each image has three kinds of textual descriptions in raw data. During pre-training, we construct an image-text pair per iteration by randomly selecting one of them. Besides, we also conduct a variety of combinations of different types of textual information, such as without ``Title''. We find the best mode is to use all three types of text instead of other combinations, which brings more data diversity and potentially improves the model performance.}


\mmljc{
\textbf{Why is \pretraindata{} large-scale and high-quality?} As shown in Table~\ref{pretrain_dataset}, the proposed \pretraindata{} is a large-scale Chinese pre-training dataset with 250 million images and 750 million corresponding texts. To the best of our knowledge, \pretraindata{} is the largest Chinese pre-training image-text dataset. Relying on the CTR filtering method and the diverse textual descriptions, \pretraindata{} is high-quality because a small amount of pre-training data (about 10\% of \pretraindata{}, 23M) surpasses the previous state-of-the-art~\cite{gu2022wukong}, which uses 100M image-text pairs. More details can be found in Section~\ref{sec:ablation}.
}




\begin{table}[t]
\centering 
\caption{\blue{Statistics of the vision-language downstream datasets. Our downstream datasets can refer to Section~\ref{downstream_dataset_intro}.}}
\resizebox{0.488\textwidth}{!}{
\begin{tabular}{lcccc} 
\toprule
\multirow{2.5}{*}{Dateset} &
\multirow{2.5}{*}{Annotation} &
\multicolumn{3}{c}{Image-Text Pairs} 
\\
\cmidrule{3-5} 
&  & Train &  Val & Test  \\
\midrule
Flickr30k-CN~\cite{flickr30k-cn} & Machine Translation & 29K  & 1K &  1K   \\
COCO-CN~\cite{coco-cn} & Human Annotation & 18K  & 1K  & 1K   \\

AIC-ICC~\cite{aic} & Human Annotation & 210K  & 30K  & 30K   \\
MUGE~\cite{M6} & - & 129K  & 29K  & 30K \\
ECommerce-T2I~\cite{M6} & - & 9K & 5K & 5K  \\
\midrule
Flickr30k-CNA & Human Annotation & 29K &  1K &  1K  \\
ICR  & Human Annotation & 160K &  20K & 20K  \\
IQR  & Human Annotation & 160K &  20K &  20K \\
ICM & Human Annotation & 320K &40K & 40K \\
IQM  & Human Annotation & 320K  & 40K &  40K  \\

\bottomrule 
\end{tabular}
}
\label{downnstream_dataset}
\end{table}

\begin{figure*}[htb]
    \centering
	\includegraphics[width=0.99\textwidth]{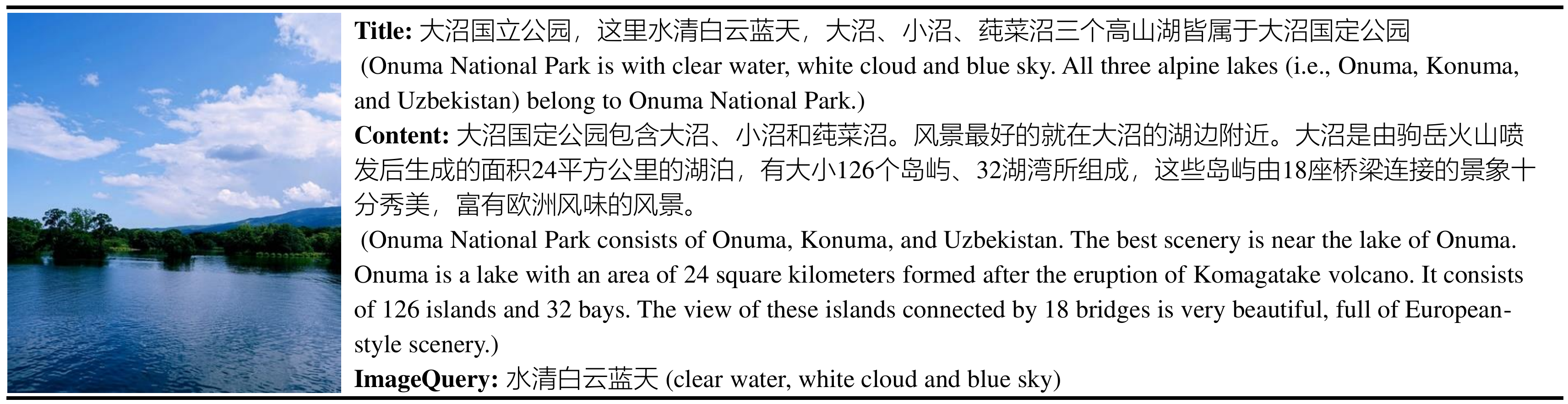}
	\caption{{An example of \mmcr{Zero}. More samples can be found in Appendix.}
	}
	\label{fig:pretraindata_one_image}
\end{figure*}

\subsection{Downstream Dataset}
\label{downstream_dataset_intro}
We perform human annotation in downstream datasets, \ie ICM, IQM, ICR, IQR, and Flickr30k-CNA. For the first four datasets, the selection strategy of image-text pairs is the same as the collection of the pre-training dataset \mmcr{Zero}. We divide the training set, validation set, and test set with a ratio of 8:1:1. We check that the downstream data do not appear in the pre-training dataset via the hash value of images. Then, 15 human annotators carefully label the image-text pairs. Specifically, the human annotators verify whether an image-text pair is relevant. That is, the human annotators mark them as positive or negative pairs until a pre-defined data size is reached. Note that we do not rewrite the caption or query for each image in these downstream datasets. For Flickr30k-CNA, we gather 6 professional English and Chinese linguists to meticulously translate all data of Flickr30k~\cite{Flickr} and double-check each sentence. The details of each dataset are as follows.


\textbf{Image-Caption Matching Dataset (ICM).}
\ljc{ICM is collected for the image-text matching task.}
\mmxie{The image-text matching task is a binary classification task, aiming to predict whether an image-text pair is matched.}
{Each image has a corresponding caption text.}
\mmljc{We first select image-text pairs beyond the 5 billion data.}
Then, human annotators manually perform a 2nd round manual correction, 
obtaining 400,000 image-text pairs, including 200,000 positive cases and 200,000 negative cases.
We keep the ratio of positive and negative pairs consistent in each of the train/val/test sets.

\textbf{Image-Query Matching Dataset (IQM).} \ljc{This is a dataset also for the image-text matching task.}
Different from ICM, we use the search query instead of detailed description text. 
Similarly, IQM contains 200,000 positive cases and 200,000 negative cases.
\mmljc{ICM and IQM are the largest Chinese vision-language downstream datasets.}

\textbf{Image-Caption Retrieval Dataset (ICR).} 
{We collect 200,000 image-text pairs under the rules described in ICM. It contains image-to-text and text-to-image retrieval tasks.}

\textbf{Image-Query Retrieval Dataset (IQR).} IQR is also proposed for the image-text retrieval task. \blue{We collect 200,000 queries} and the corresponding images as the annotated image-query pairs similar to IQM. 
\mmxie{We show examples of the above four datasets in \mmcr{Figure B in Appendix.}}

\textbf{Flickr30k-CNA Dataset.}
Former Flickr30k-CN~\cite{flickr30k-cn} translates the training and validation sets of Flickr30k~\cite{Flickr} using machine translation, and manually translates the test set.
We check the machine-translated results and find two kinds of problems. (1) Some sentences have language problems and translation errors. (2) Some sentences have poor semantics. 
In addition, the different translation ways prevent the model from achieving accurate performance.
We gather 6 professional English and Chinese linguists to meticulously re-translate all data of Flickr30k and double-check each sentence.
We name this dataset as Flickr30k-Chinese All (Flickr30k-CNA). 
We show some cases of the difference between Flickr30k-CN and Flickr30k-CNA in Appendix.

\section{Methodology}\label{approach}

\subsection{Model Architecture}\label{sec model}
\begin{figure*}[t]
    \centering
	\includegraphics[width=0.99\textwidth]{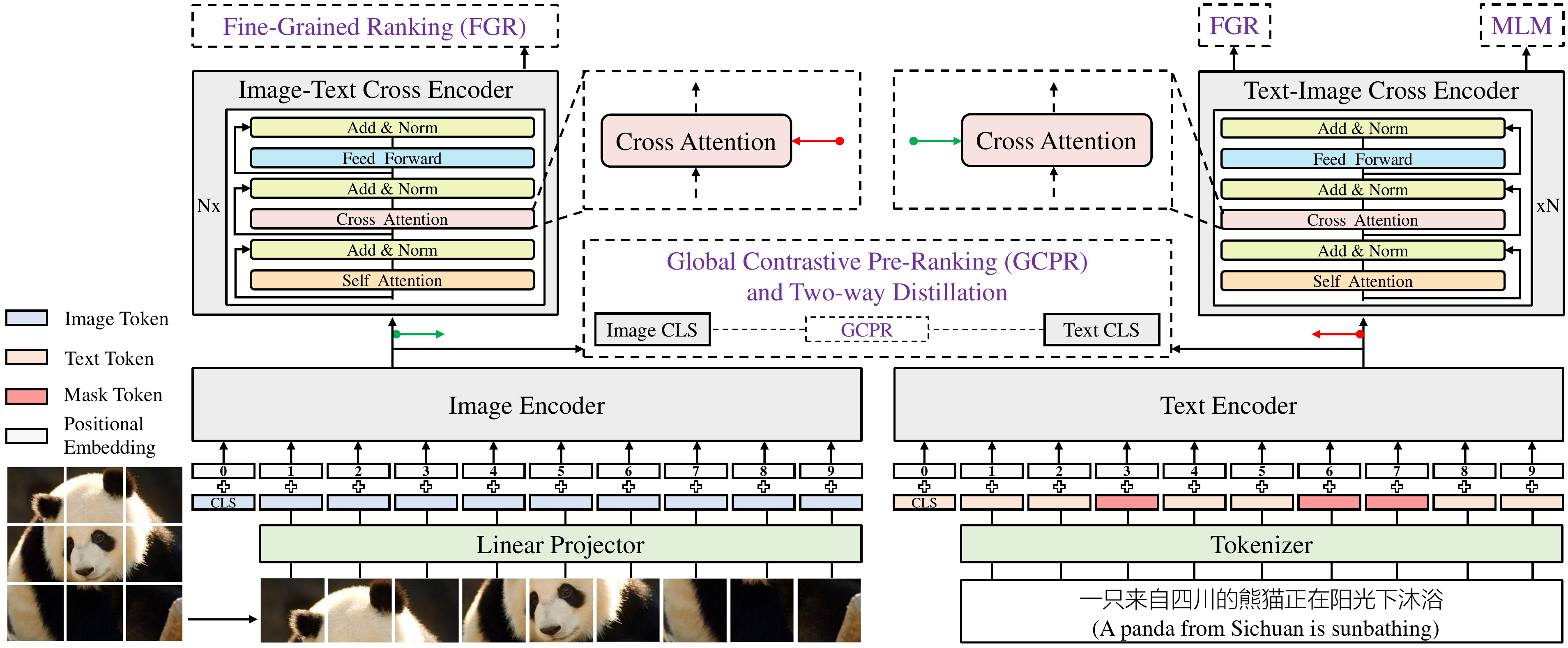}
	\caption{The overall architecture of the proposed framework. The image encoder and the text encoder aim to learn individual features of image and text, respectively. Then, the image features (green circled arrow) are fed into the text-image cross encoder. Similarly, the text features (red circled arrow) are fed into the image-text cross encoder. 
    \mmcr{During pre-training, we apply global contrastive pre-ranking (GCPR), fine-grained ranking (FGR), two-way distillation (TwD), and mask language modeling (MLM) with enhanced training (ET) as pre-training objectives.}
    }
	\label{fig:framework}
\end{figure*}

From Figure \ref{fig:framework}, \blue{the architecture} contains a text encoder, an image encoder, and two cross encoders.
\blue{The text encoder is a BERT~\cite{bert} using the tokenizer of RoBERTa-wwm-ext~\cite{chinesebert}. For the image encoder, we adopt the Vision Transformer (ViT)~\cite{vit}. The two cross encoders are multi-layer transformers.}
The text encoder and image encoder transform texts and images into sequences of hidden states separately.
Then the text and image hidden states interact in the two cross encoders through cross-attention.

\subsection{Pre-training Objectives}\label{sec Objectives}
\mmxie{We jointly optimize \mmcr{R2D2} with \mmljc{the following} four objectives.}
\mmcr{To fully explore the matching relationship between image and text pairs, we design a mechanism of pre-ranking + ranking, named global contrastive pre-ranking (GCPR) and fine-grained ranking (FGR). To further enhance the capability of the model, we propose a two-way distillation (TwD) strategy consisting of target-guided distillation and feature-guided distillation. We adopt masked language modeling (MLM) with enhanced training (ET) to effciently learn the representation of cross-modal models.}
We conduct the ablation study to verify the \mmljc{effectiveness} of each pre-training strategy in Section \ref{sec:ablation}.

\textbf{Global Contrastive Pre-Ranking.}
\mmxie{Our global contrastive pre-ranking method is similar to that of CLIP~\cite{CLIP}, aiming to align the representation of multi-modal data (\eg paired image and text). 
The open-source CLIP implementation~\cite{CLIP} only performs back-propagation of the contrastive loss from the local GPU, where negative samples are not fully utilized. Instead, We back-propagate the gradients across all $k$ GPUs. Inspired by MoCo~\cite{moco}, we also introduce a queue mechanism. In practice, two queues with a fixed size $M$ aim to maintain the recent image and text representations from the momentum-updated encoders, respectively.  
For each image $I_{i}$ and the corresponding text $T_{i}$, the softmax-normalized similarity score of image-to-text and text-to-image can be defined as:}

\begin{equation}
    \begin{aligned}
    \label{IT-score-multi-gpu}
    \bs(I_{i}, T_{i}) = \frac{\exp(\simm(I_{i}, T_{i})/\tau)}{\sum_{j=1}^{n\times k+M} \exp(\simm(I_{i}, T_{j})/\tau)}, \\
    \bs(T_{i}, I_{i}) = \frac{\exp(\simm(T_{i}, I_{i})/\tau)}{\sum_{j=1}^{n\times k+M} \exp(\simm(T_{i}, I_{j})/\tau)}, 
    \end{aligned}
\end{equation}

\ljc{where $n$ is the batch size of one GPU, $k$ is the number of GPUs, $\tau$ is a learnable temperature parameter, and $\simm(\cdot, \cdot)$ denotes the cosine similarity between a pair of image-text.}
\mmxie{Considering the effectiveness of features in the queue decreases with increasing time steps, we also maintain a weighted queue $w$ to mark the reliability of the corresponding position features. Specifically, we decay each element in the queue by a factor of 0.99 per iteration, except for the new incoming item. Let $\mD$ denote the training data and $\by(\cdot, \cdot)$ denote the ground-truth one-hot label. The \mmcr{global contrastive pre-ranking (GCPR)} loss is calculated by the weighted cross-entropy loss $\mL_w(\cdot)$, as shown in Equation (\ref{ITCloss_gcpr}). }

\begin{equation*}
    \begin{aligned}
        \mL_{i2t}^{w}(I,T)=\mL_w(\bs(I, T), \by(I, T);w), \\
        \mL_{t2i}^{w}(T,I)=\mL_w(\bs(T, I), \by(T, I);w),
    \end{aligned}
\end{equation*}
\begin{equation}
    \begin{aligned}
    \label{ITCloss_gcpr}
    \mL_{\text{GCPR}} = \frac{1}{2}\mmE_{(I,T){\sim} \mD} \!\left [\mL_{i2t}^{w}(I,T) + \mL_{t2i}^{w}(T,I) \right ].
    \end{aligned}
\end{equation}

\textbf{Fine-Grained Ranking.}
As aforementioned, we apply \mmcr{global contrastive pre-ranking} to obtain the individual representations of images and texts, respectively. Relying on these representations, we next perform \mmcr{Fine-Grained Ranking (FGR)} loss. To be specific, this is a binary classification task, aiming to predict whether an image-text pair is matched.
Formally, we denote $h_{I_{[CLS]}}$ and $h_{T_{[CLS]}}$ as the output representations of two cross encoders.
Given an image representation $h_{I_{[CLS]}}$ and a text representation $h_{T_{[CLS]}}$, 
we feed the representations into a fully-connected layer $g(\cdot)$ to get the predicted probabilities respectively. Let $\by$ denote the ground-truth label of binary classification, we then compute the \mmcr{FGR} loss by the cross-entropy loss $\mL_c(\cdot)$ as:
\begin{align}
    \label{FGRloss}
    \!\!\!\mL_{\text{FGR}}\!\!=\!\!\frac{1}{2}\!\mathop{\mmE}_{(I,T){\sim} \mD}\! \!\left [\mL_c(g(h_{I_{[CLS]}}),\!\by)\!\!+\!\!\mL_c(g(h_{T_{[CLS]}}),\! \by)\right ]\!
\end{align}

\blue{The selection strategy of negative pairs is in Appendix.}

\textbf{Two-way Distillation.}
\mmljc{Relying on the momentum-updated encoders in contrastive learning, we introduce target-guided distillation (TgD) to decrease the risk of learning from noisy labels, and feature-guided distillation (FgD) to improve the generalization performance of the pre-trained model. We conduct target-guided distillation to learn from pseudo-targets generated by the momentum model following ALBEF~\cite{ALBEF}. In practice, we replace the target in Equation (\ref{ITCloss_gcpr}) with the pseudo-targets. \mmcr{More details about the training process of TgD can be found in Appendix.}
Besides, target-guided distillation and feature-guided distillation both adopt a teacher-student paradigm.}
\mmljc{For convenience, we call the combination of TgD and FgD as two-way distillation (TwD). }


\mmljc{Below are the details of FgD.}
Taking the text encoder as the example below, the teacher character is the momentum-updated text encoder and the student is the text encoder. Here, the weights of the teacher are updated by all past text encoders via exponential-moving-average. 
To further improve the capability of the model, we apply a masking strategy to the inputs. In practice, we feed complete inputs into the teacher and masked inputs into the student. 
Relying on the momentum mechanism, we aim to make the features of the student closer to that of the teacher.
Formally, the predicted distributions (\ie $\mP_t(T)$, $\mP_s(T)$) of the teacher and the student are defined as follows, 
respectively.
\begin{equation}
    \begin{aligned}
    \label{FgD_Eqn}
    \mP_t(T) = \frac{\text{exp}((f_t(T)-\mu)/\tau_t)}{\sum_{i=1}^d \text{exp}((f_t(T)^{(i)}-\mu^{(i)})/\tau_t)}, \\
    \!\!\!\!\!\!\!\!\!\!\!\!\!\!\!\!\!\!\!\mP_s(T) = \frac{\text{exp}(f_s(T)/\tau_s)}{\sum_{i=1}^d \text{exp}(f_s(T)^{(i)}/\tau_s)},
    \end{aligned}
\end{equation}
where $f_t(\cdot)$ and $f_s(\cdot)$ denote the networks of the teacher and the student, respectively. \xie{Moreover, $\mu$ is a momentum-updated mean of $f_t(\cdot)$, and $d$ is the dimension of the features. $\tau_t$ and $\tau_s$ are the temperature parameters of the teacher and the student, respectively, which can
sharpen the distribution of the features.}
Note that we do not use $\mu$ for $\mP_s$ to avoid collapse in feature-guided distillation. We can obtain similar formulations for $\mP_s(I)$ and $\mP_t(I)$. We perform the feature-guided distillation by the cross-entropy loss, and the loss $L_{\text{FgD}}$ is defined as:
\begin{align}
    \!\mL_{\text{FgD}}\!\!=\!\!\frac{1}{2}\!\mathop{\mmE}_{(I,T){\sim} \mD} \![\mL_c(\mP_s(I),\!\mP_t(I))\!\!+\!\!\mL_c(\mP_s(T),\!\mP_t(T))].\!
\end{align}
\mmxie{Through experiments in Section~\ref{sec:ablation}, we observe
a noticeable performance gain by performing FgD.}

\textbf{\blue{Masked Language Modeling with Enhanced Training}.}
We apply a masked language modeling loss to the text-image cross encoder to improve the ability to model the relationship between image and text at the token level. 15\% of the text tokens are masked in the input. All of these tokens are replaced with the $[MASK]$ token.
\mmxie{The forward operations of MLM~\cite{bert} and \mmcr{FGR} are executed individually in most VLP models~\cite{uniter,ALBEF,FLAVA}}, increasing the computational cost of pre-training. In our model, the MLM task utilizes masked text and corresponding images together for denoising, which enhances the interaction between text and images. \mmxie{Since \mmcr{FGR} relies heavily on this interaction ability, we propose enhanced training (ET), applying \mmcr{FGR} and MLM loss on text tokens with masking simultaneously.}
Experiments in Section \ref{sec:ablation} show that ET can reduce the computational cost of \mmcr{R2D2} while maintaining the accuracy of the model. For simplicity, $\mL_{\text{MLM}}$ denotes the loss of the MLM task \blue{with enhanced training}.
\mmxie{Our model is trained with the full objective:
\begin{align}
    \mL = \mL_{\text{GCPR}}^{\text{TgD}}+\mL_{\text{FgD}}+\mL_{\text{\mmcr{FGR}}}+\mL_{\text{MLM}}.
\end{align}
}

\section{Experiments}
\subsection{Implementation Details}\label{sec:implementation}
The number of transformer layers for the text encoder, and the two cross encoders are 12, 6, and 6, respectively. 
The text encoder is initialized from RoBERTa-wwm-ext~\cite{chinesebert} while the two cross encoders are randomly initialized.
Following Wukong~\cite{gu2022wukong}, we use the image encoder of 12-layers ViT-Base and 24-layers ViT-Large initialized from CLIP~\cite{CLIP}, and freeze it during pre-training. 
The resolution of the input image is 224$ \times $224 in pre-training and fine-tuning. The dimension of the feature vectors of both image and text is 768. \blue{We pre-train models with 15 epochs using a batch size of 4096 on 128 NVIDIA A100 GPUs.}
\mmcr{$\tau$ in Equation \ref{IT-score-multi-gpu} is initialized to 0.07 and can be learned during the training process.}
We set $\tau_s$= 0.1 and $\tau_t$= 0.04 in Equation \ref{FgD_Eqn}.
Moreover, the momentum is set as $m= 0.995$, and the queue size is 36,864.
\blue{We adopt the Adam optimizer and the cosine learning rate schedule with a linear warmup~\cite{loshchilov2016sgdr}.}
\blue{The pre-trained model is adapted to five vision-language downstream tasks: image-text retrieval, image-text matching, image caption, text-to-image generation, and zero-shot image classification.}
\mmljc{More details can refer to Appendix.}

\begin{table}[t]
    \caption{Comparisons with state-of-the-art models on image-text retrieval task. \blue{CNA represents our Flickr30k-CNA.}}
  \centering
  \resizebox{0.48\textwidth}{!}{
  \begin{tabular}{c|lccccccc}
    \toprule
    &
    \multirow{2.5}{*}{Method} &
    \multicolumn{3}{c}{Image-to-Text Retrieval} &
    \multicolumn{3}{c}{Text-to-Image Retrieval} &
    \multirow{2.5}{*}{R@M}
    \\ 
    \cmidrule(r){3-5} \cmidrule(r){6-8}
    & &  R@1 & R@5 & R@10 & R@1 & R@5 & R@10
    \\
    \midrule
    \multirow{8}*{\rotatebox{90}{Flickr30k-CN}} 
    & $\text{CLIP}_\text{ViT-B}$ & 87.1 & 97.7 & 98.8 &69.0 &90.3 &95.0 &89.7 \\
    
    & $\text{CLIP}_\text{ViT-L}$~\cite{CLIP}  & 91.6 & 99.1 & 99.7 & 77.3 & 94.4 & 97.2 & 93.2 \\
    
    & $\text{FILIP}_\text{ViT-B}$ &72.1  &91.3 &95.8 &57.5 &84.3 &90.6 &81.9 \\
    & $\text{FILIP}_\text{ViT-L}$~\cite{filip}  & 90.6 & 98.8 & 99.6 & 76.9 & 94.9 & 97.4 & 93.0 \\
    & $\text{Wukong}_\text{ViT-B}$  & 83.9 & 97.6 & 99.0 & 67.6 & 89.6 & 94.2 & 88.7 \\
    & $\text{Wukong}_\text{ViT-L}$~\cite{gu2022wukong}  & 92.7 & 99.1 & 99.6 & 77.4 & 94.5 & 97.0 & 93.4 \\
    & $\text{\mmcr{R2D2}}_\text{ViT-B}$ & 93.2 & 99.2 & 99.8 & 79.2 & 95.2 & 97.3 & 94.0  \\
    & $\text{\mmcr{R2D2}}_\text{ViT-L}$&  \textbf{95.6} & \textbf{99.8} & \textbf{100.0} & \textbf{84.4} & \textbf{96.7} & \textbf{98.4} & \textbf{95.8}  \\
    \midrule

    \multirow{8}*{\rotatebox{90}{COCO-CN}}
    & $\text{CLIP}_\text{ViT-B}$ & 68.7 & 93.6 & 97.5 &68.9 &93.3 &97.3 &86.6 \\
    & $\text{CLIP}_\text{ViT-L}$~\cite{CLIP}  & 68.3 &93.0 &97.3 &70.1 &92.2 &96.4 &86.2 \\
    & $\text{FILIP}_\text{ViT-B}$ &52.7  &81.3 &88.3 &56.2 &86.8 &94.3 &76.6 \\
    & $\text{FILIP}_\text{ViT-L}$~\cite{filip}  & 69.1 &91.3 &96.9 &72.2 &92.4 &97.2 &86.5 \\
    & $\text{Wukong}_\text{ViT-B}$  & 65.8 & 90.3 & 96.6 & 67.0 & 91.4 & 96.7 & 84.6 \\
    & $\text{Wukong}_\text{ViT-L}$~\cite{gu2022wukong} & 73.3 & 94.0 & 98.0 & 74.0 & 94.4&  98.1 & 88.6 \\
    & $\text{\mmcr{R2D2}}_\text{ViT-B}$ &  78.1 & 96.2 & 98.6 & 76.0 & 94.9 & 98.3 & 90.3 \\
    & $\text{\mmcr{R2D2}}_\text{ViT-L}$ & \textbf{79.3} & \textbf{97.1} & \textbf{98.7} & \textbf{79.1} & \textbf{96.5} & \textbf{98.9} & \textbf{91.6} \\
    \midrule
    \multirow{11}*{\rotatebox{90}{AIC-ICC}}
    & BriVL~\cite{wenlan2} &  45.6 & 68.0 & 76.3 & 34.1 & 58.9 & 69.1 & 58.7 \\
    & $\text{CLIP}_\text{ViT-B}$ & 50.5 &73.0 &80.2 &38.1 &63.7 &73.3 &63.1 \\
    & $\text{CLIP}_\text{ViT-L}$~\cite{CLIP}  &59.1 &79.5 &85.2 &46.2 &70.7 &78.6 &69.9\\
    & $\text{FILIP}_\text{ViT-B}$ &42.5  &67.2 &76.0 &32.9 &58.4 &68.8 &57.6 \\
    & $\text{FILIP}_\text{ViT-L}$~\cite{filip}  &54.1 &75.8 &82.8 &44.9 &69.0 &77.5 &67.4 \\
    & $\text{Wukong}_\text{ViT-B}$ & 47.5 & 70.6 & 78.6 & 36.7 & 36.7 & 71.7 & 57.0 \\
    & $\text{Wukong}_\text{ViT-L}$~\cite{gu2022wukong} &61.6 & \textbf{80.5} & \textbf{86.1} & 48.6 & 72.5 & 80.2 & 71.6 \\
    & $\text{\mmcr{R2D2}}_\text{ViT-B}$&  56.8& 76.2& 82.1& 47.6 & 72.8 & 80.2 & 69.3 \\
    & $\text{\mmcr{R2D2}}_\text{ViT-L}$& \textbf{65.4} &  80.3 &  84.7 & \textbf{57.3} & \textbf{78.1} & \textbf{83.0} & \textbf{74.8} \\
    \midrule

    \multirow{8}*{\rotatebox{90}{MUGE}}
    & $\text{CLIP}_\text{ViT-B}$ &-  &- &- &43.5 &71.7 &80.6 &65.3 \\
    & $\text{CLIP}_\text{ViT-L}$~\cite{CLIP}  & - & - & - & 50.1 &76.9 &84.9 &70.6 \\
    & $\text{FILIP}_\text{ViT-B}$ &-  &- &- &30.6 &58.2 &70.2 &53.0 \\
    & $\text{FILIP}_\text{ViT-L}$~\cite{filip} & - & - & - & 43.5 &71.5 &80.9 &65.3 \\
    & $\text{Wukong}_\text{ViT-B}$ & - & - & - & 39.2 & 66.9 & 77.4 & 61.2 \\
    & $\text{Wukong}_\text{ViT-L}$~\cite{gu2022wukong} &- & - & - & 52.7 & 77.9 & 85.6 & 72.1\\
    & $\text{\mmcr{R2D2}}_\text{ViT-B}$& - & - & - & 53.4 & 78.1 & 86.0 & 72.5  \\
    & $\text{\mmcr{R2D2}}_\text{ViT-L}$&  - & - & - & \textbf{60.1} & \textbf{82.9} & \textbf{89.4} & \textbf{77.5} \\
    \midrule

    \multirow{2}*{\rotatebox{90}{CNA}}
    & $\text{\mmcr{R2D2}}_\text{ViT-B}$  & 93.6 & 99.5 & 99.8 & 80.5 & 95.6 & 97.7 & 94.5 \\
    & $\text{\mmcr{R2D2}}_\text{ViT-L}$ & \textbf{96.9} & \textbf{99.8} & \textbf{100.0} & \textbf{84.9} & \textbf{97.0} & \textbf{98.6} & \textbf{96.2}\\
    \midrule

    \multirow{2}*{\rotatebox{90}{ICR}}
    & $\text{\mmcr{R2D2}}_\text{ViT-B}$& 53.4 & 75.4 & 83.4 & 52.1 & 73.3 & 82.0 & 69.9   \\
    & $\text{\mmcr{R2D2}}_\text{ViT-L}$& \textbf{61.5} & \textbf{82.9} & \textbf{87.7} & \textbf{60.7} & \textbf{82.0} & \textbf{86.9} & \textbf{77.0} \\
    \midrule

    \multirow{2}*{\rotatebox{90}{IQR}}
    & $\text{\mmcr{R2D2}}_\text{ViT-B}$& 37.0 & 62.1 & 70.9 & 35.8 & 61.2 & 70.5 & 56.3  \\
    & $\text{\mmcr{R2D2}}_\text{ViT-L}$&  \textbf{41.9} & \textbf{67.8} & \textbf{75.9} & \textbf{41.3} & \textbf{67.6} & \textbf{75.4} & \textbf{61.7} \\
    \bottomrule
  \end{tabular}
  }
  \label{itr-sota}
\end{table}

\begin{table*}[h]
  \caption{Comparison with state-of-the-art models on downstream vision-language tasks.}
  \centering
  \resizebox{0.99\textwidth}{!}{
  \begin{tabular}{lcccccccccc}
    \toprule
    \multirow{2}{*}{Method} &
    \multicolumn{2}{c}{Image-Text Matching} &
    \multicolumn{4}{c}{Image Caption} &
    \multicolumn{2}{c}{Text-to-Image Genearation} &
    \multicolumn{2}{c}{Zero-shot Image Classification}
    \\ 
    \cmidrule(r){2-3} \cmidrule(r){4-7} \cmidrule(r){8-9} \cmidrule(r){10-11}
    & AUC (ICM) & AUC (IQM) & BLEU & METEOR & ROUGE-L &	CIDEr & \multicolumn{2}{c}{FID} &  \quad Top-1 Acc. & Top-5 Acc.
    \\
    \midrule
    BriVL~\cite{wenlan2} &61.9  &57.6  &66.1  &41.1  &71.9  &220.7  & \multicolumn{2}{c}{-}  & \quad 24.3  &  56.8 \\
    $\text{Wukong}_\text{ViT-B}$ &79.2  &75.1  &66.7  &71.2  &72.2  &224.2  & \multicolumn{2}{c}{23.7} & \quad 49.1  & 74.2 \\
    $\text{Wukong}_\text{ViT-L}$~\cite{gu2022wukong} &81.8  &78.1  &68.9  &74.5  &72.3  &243.1  & \multicolumn{2}{c}{18.8} & \quad 55.0  &80.5 \\
    \midrule
    $\text{\mmcr{R2D2}}_\text{ViT-B}$ & 88.6 & 84.9 & 68.3 & 76.3 & 73.2 & 230.2 &  \multicolumn{2}{c}{18.9}  &\quad 50.6 & 78.1\\
    $\text{\mmcr{R2D2}}_\text{ViT-L}$ & \textbf{90.6} & \textbf{86.7} & \textbf{71.8} & \textbf{78.2} & \textbf{75.3} & \textbf{247.9} &  \multicolumn{2}{c}{\textbf{14.4}} &\quad \textbf{56.9} &\textbf{83.3} \\
    \bottomrule
  \end{tabular}
  }
  \label{Downstream:table}
\end{table*}

\begin{table*}[h]
\caption{\mmxie{
  Comparison with state-of-the-arts which combine dual-stream and single-stream architectures. Classification represents zero-shot image classification. We report R@M, AUC, CIDEr, FID, and Top-1 accuracy for five V+L downstream tasks respectively.}}
  \centering
  \resizebox{0.89\textwidth}{!}{
  \begin{tabular}{lcccccccc}
            \toprule
            \multirow{2}{*}{Method} &
            \multicolumn{2}{c}{Image-Text Retrieval} &
            \multicolumn{2}{c}{Image-Text Matching} &
            \multicolumn{1}{c}{Image Caption} &
            \multicolumn{1}{c}{Text-to-Image Generation}&
            \multicolumn{1}{c}{Classification}
            \\ 
            \cmidrule(r){2-3} \cmidrule(r){4-5} \cmidrule(r){6-6} \cmidrule(r){7-7} \cmidrule(r){8-8}
             & Flick30k-CN & COCO-CN & \quad ICM & IQM &  AIC-ICC &  ECommerce-T2I & ImageNet\\
            \midrule
            ALBEF\cite{ALBEF}  &  90.1 & 84.9 & \quad79.5 & 74.7 & 226.0 & 21.4 &35.9 \\
            FLAVA\cite{FLAVA} &  91.4 & 85.1 & \quad80.1 & 75.6 &  226.2 &  21.0 &37.2 \\
            \midrule
            $\text{\mmcr{R2D2}}_\text{ViT-B}$ &  \textbf{92.2} & \textbf{86.3} & \quad\textbf{81.1} & \textbf{76.3} &  \textbf{226.8} &  \textbf{20.9} &\textbf{37.5}\\
            
            \bottomrule
          \end{tabular}
}
  \label{Downstream:albef-flava}
\end{table*}

\begin{table*}[h]
\caption{\mmxie{
  Effect of the proposed pre-training dataset. Classification represents zero-shot image classification.  We report R@M, AUC, CIDEr, FID, and Top-1 accuracy for five V+L downstream tasks respectively.}}
  \centering
  \resizebox{0.99\textwidth}{!}{
  \begin{tabular}{lcccccccc}
            \toprule
            \multirow{2}{*}{Method} &
            \multirow{2}{*}{Pre-training Dataset} &
            \multicolumn{2}{c}{Image-Text Retrieval} &
            \multicolumn{2}{c}{Image-Text Matching} &
            \multicolumn{1}{c}{Image Caption} &
            \multicolumn{1}{c}{Text-to-Image Generation}&
            \multicolumn{1}{c}{Classification}
            \\ 
            \cmidrule(r){3-4} \cmidrule(r){5-6} \cmidrule(r){7-7} \cmidrule(r){8-8} \cmidrule(r){9-9}
             & & Flick30k-CN & COCO-CN & \quad ICM & IQM &  AIC-ICC &  ECommerce-T2I & ImageNet\\
            \midrule
            \mmcr{R2D2} & Wukong (100M)~\cite{gu2022wukong} &  95.2 & 90.1 & \quad86.5 & 81.5 & 245.8 & 16.4 &55.6 \\
            \midrule
            \mmcr{R2D2} & \mmcr{Zero} (23M) &  95.4 & 90.7 & \quad88.1 & 83.6 &  246.5 &  15.7 &55.7 \\
            \mmcr{R2D2} & \mmcr{Zero} (250M) &  \textbf{95.8} & \textbf{91.6} & \quad\textbf{90.6} & \textbf{86.7} &  \textbf{247.9} &  \textbf{14.4} &\textbf{56.9}\\
            
            \bottomrule
          \end{tabular}
}
  \label{Downstream:tablex3}
\end{table*}

\begin{table*}[!h]
\caption{Effect of different components of \pretrainmodel{}. Note that we conduct ablation studies and report the average results on all downstream datasets.  \blue{Generation and classification represent text-to-image generation and zero-shot image classification, respectively. R@* denotes the result for the image-text retrieval task. We report AUC, CIDEr, FID, and Top-1 accuracy for image-text matching, image caption, text-to-image generation, and zero-shot image classification tasks respectively.}}
  \centering
  \resizebox{0.99\textwidth}{!}{
  \begin{tabular}{lccccccccccc}
    \toprule
    \multirow{2}{*}{Method} &
    \multicolumn{3}{c}{Image-to-Text Retrieval} &
    \multicolumn{3}{c}{Text-to-Image Retrieval} &
    \multirow{2}{*}{R@M} &
    \multicolumn{1}{c}{Image-Text Matching} &
    \multicolumn{1}{c}{Image Caption} &
    \multicolumn{1}{c}{Generation} &
    \multicolumn{1}{c}{Classification}
    \\ 
    \cmidrule(r){2-4} \cmidrule(r){5-7} \cmidrule(r){9-9} \cmidrule(r){10-10} \cmidrule(r){11-11} \cmidrule(r){12-12}
    & R@1 & R@5 & R@10 & R@1 & R@5 & R@10 &  & AUC &CIDEr &FID  &Top-1 Acc. 
    \\
    \midrule
    \mmcr{PRD2} &53.92  &75.67  &82.01  &43.97  &71.19  &80.28  &67.14  & 73.89 &239.91 &19.21 &32.27\\
    \midrule
    \pretrainmodel{} & \textbf{64.51} & \textbf{81.02} & \textbf{85.92} & \textbf{56.63} & \textbf{78.22} & \textbf{84.49} & \textbf{74.45} & \textbf{80.82} &\textbf{243.29} &\textbf{17.58} &\textbf{39.96}\\
	\pretrainmodel{} w/o ET & 64.14 & 78.48 & 84.96 & 55.32 & 77.38 & 83.01 & 73.81 & 80.31 & 243.01 &17.82 &39.72\\
 	\pretrainmodel{} w/o MLM & 63.72 & 80.19 & 85.14 & 55.73 & 77.29 & 83.70 & 73.57 & 80.01 &242.90 &17.91 &39.54\\
	\pretrainmodel{} w/o TwD & 63.08 & 79.51 & 84.69 & 54.69 & 76.74 & 83.53 & 73.03  & 79.98 &242.64 &18.16 &38.76   \\
	\pretrainmodel{} w/o TgD & 63.87 & 80.43 & 85.39 & 55.97 & 77.02 & 83.23 & 73.52 & 80.39 &243.01 &17.90 &39.29  \\
	\pretrainmodel{} w/o FgD & 63.39 & 79.86 & 85.01 & 54.92 & 76.83 & 83.45 & 73.11 & 80.28  &242.83 &18.01 &38.92  \\
    \bottomrule
  \end{tabular}
  }
  \label{Ablation:table}
\end{table*}

\subsection{Comparisons with State-of-the-art}\label{sec:SOTA}
\xie{For both image-to-text retrieval and text-to-image retrieval tasks, we report Recall@1 (R@1), Recall@5 (R@5), Recall@10 (R@10), and Mean Recall (R@M).
The results of BriVL~\cite{wenlan2} and Wukong~\cite{gu2022wukong} are excerpted from their paper. Wukong reproduces the CLIP-style~\cite{CLIP} and FILIP-style~\cite{filip} models.
\ldw{Their results are also included.}
From Table \ref{itr-sota}, our models outperform state-of-the-art on all datasets.
Moreover, $\text{\mmcr{R2D2}}_\text{ViT-L}$ outperforms $\text{\mmcr{R2D2}}_\text{ViT-B}$.}
\jc{These results indicate that our framework is able to learn better fine-grained associations between image and text.}
\jc{We report the results of Flickr30k-CNA on the test set of Flickr30k-CN for a fair comparison.} \mmcr{R2D2} fine-tuned on Flickr30k-CNA outperforms that on Flickr30k-CN, since the quality of human-translated Flickr30k-CNA is much higher than that of machine-translated Flickr30k-CN.

Table \ref{Downstream:table} reports the comparison with existing methods on other V+L downstream tasks. Unlike the image-text retrieval task, there are few datasets for the Chinese image-text matching (ITM) task. Thus, we introduce image-caption matching dataset (ICM) and image-query matching dataset (IQM) for the Chinese ITM task and show the corresponding results. 
Also, we evaluate Wukong and BriVL on these datasets for the ITM task.
We use Area Under Curve (AUC) as the metric. For the image captioning task, fine-tuning is conducted on the training split of AIC-ICC~\cite{aic}. We adopt four widely-used evaluation metrics: BLEU, METEOR, ROUGE-L, and CIDEr following BriVL. Table \ref{Downstream:table} also presents text-to-image generation results on \blue{ECommerce-T2I dataset\footnote{
https://tianchi.aliyun.com/muge} \cite{M6}}. The metric of Frechet Inception Distance (FID) is reported. \blue{We evaluate our pre-trained models on ImageNet\cite{imagenet} for the zero-shot image classification task. Class labels are translated from English. Top-1 and Top-5 accuracy are reported.}
Our model achieves state-of-the-art performance on these V+L downstream tasks. 

\mmxie{ALBEF~\cite{ALBEF} and FLAVA~\cite{FLAVA} also combine dual-stream unimodal encoders and single-stream multimodal encoders. They are pre-trained with English Corpus, lacking the ability to perform Chinese downstream tasks. To make a comparison with these methods, We pre-train ALBEF, FLAVA, and \mmcr{R2D2} on the first 1\% of \mmcr{Zero}. We replace the text encoder and tokenizer of these baselines with the same as ours. Considering that ALBEF and FLAVA use ViT-Base as the image encoder, we show the comparative performance of $\text{\mmcr{R2D2}}_\text{ViT-B}$ in Table \ref{Downstream:albef-flava}. In summary, the results on various tasks \mmljc{demonstrate the superiority} of our framework.
}

\subsection{Ablation Study
}\label{sec:ablation}
\textbf{Effect of the Proposed Pre-training Dataset.} 
\mmxie{To demonstrate the effectiveness of our proposed pre-training dataset, we provide comparison results of our \mmcr{R2D2} framework pre-trained on the 100M Wukong dataset~\cite{gu2022wukong} and the proposed \mmcr{Zero} in Table~\ref{Downstream:tablex3}. \mmxie{Wukong is the \mmljc{previous} largest publicly available Chinese image-text pre-training dataset.} For simplicity, we define $\text{\mmcr{R2D2}}_\text{ViT-L}$ as \mmcr{R2D2} in the ablation study.}
\mmxie{\mmcr{R2D2} pre-trained on the 23M pre-training dataset (a subset of \mmcr{Zero}) achieves better results than the ones on the much larger 100M Wukong dataset. 
This improvement verifies the high quality of our \mmcr{Zero} dataset, which is filtered by user click-through rate and provides diverse text descriptions along with each image, compared to previous datasets.} 
\blue{Moreover, we achieve the best results on the whole pre-training dataset, \ie \mmcr{Zero} with 250M high-quality image-text pairs.
}

\mmxie{\textbf{Effect of Fine-Grained Ranking (\mmcr{FGR}).}}
\xie{We conduct subsequent ablation studies on the first \blue{1\%} of \pretraindata{}.}
\mmxie{We first train a restricted version of \mmcr{R2D2} using only the \mmcr{global contrastive pre-ranking} and the two-way distillation strategy.
We denote it as \mmcr{PRD2}.} This restricted setting is conceptually similar to CLIP~\cite{CLIP}.
\mmxie{\mmcr{R2D2} outperforms \mmcr{PRD2} on the downstream tasks, indicating that the two cross encoders can effectively interact with image and text information through cross-attention.
}

\ljc{
\textbf{Effect of Enhanced Training (ET).}
From the third row of Table \ref{Ablation:table},
\blue{\mmcr{R2D2} (with ET) performs slightly better than \mmcr{R2D2} w/o ET.} 
\blue{Furthermore,} \mmcr{R2D2} uses less computational resources than \mmcr{R2D2} w/o ET. \mmcr{R2D2} requires 154.0 GFLOPs and can run at 1.4 iterations per second (Iter/s), while without ET we get 168.8 GFLOPS and 1.1 Iter/s.}
This indicates that ET is able to both reduce the computational cost and improve the capability of the learning process.

\xie{
\textbf{Effect of Masked Language Modeling (MLM).}
Compared to \mmcr{R2D2} w/o MLM,
\mmcr{R2D2} obtains better performance on all downstream tasks.
{MLM allows \mmcr{R2D2} to learn robust representations by masking data.}
These results indicate that MLM is indeed effective for downstream tasks.
}

\textbf{Effect of Two-way Distillation (TwD).} The proposed two-way distillation is composed of target-guided distillation (TgD) and feature-guided distillation (FgD). 
By analyzing the two components of TwD, we see that performing feature alignment is important, since the model w/o FgD shows a more noticeable drop in performance.
Although milder, removing TgD also causes a reduction in performance.
These results indicate that both components are relevant and TwD is an effective way to improve the generalization performance of the pre-trained model.

\subsection{Further Experiments}
\label{sec:further_exp}
\textbf{Zero-shot Tasks.} 
\blue{In this section,}
we conduct zero-shot transfer experiments.
From Table \ref{zero-shot},
\blue{our $\text{\mmcr{R2D2}}_\text{ViT-L}$ achieves the best performance on Flickr30k-CN, COCO-CN, MUGE, AIC-ICC, ICR, and IQR.}
For example, $\text{\mmcr{R2D2}}_\text{ViT-L}$ achieves 80.5\% R@M on COCO-CN, an absolute 5.3\% gain over the previous best performance. 
These results demonstrate sound generalization ability of \mmcr{R2D2}.
The results of $\text{\mmcr{R2D2}}_\text{ViT-L}$ on Flickr30k-CNA are the same as that of Flickr30k-CN, since we use the same test set for a fair comparison.
In this way, we do not report the results of $\text{\mmcr{R2D2}}_\text{ViT-L}$ on Flickr30k-CNA.
\blue{In addition, the AUC scores of $\text{\mmcr{R2D2}}_\text{ViT-L}$ on ICM and IQM are 89.8\% and 84.5\%, respectively.}

\textbf{Entity-conditioned Image Visualization.} In this experiment, we visualize the attention map of images on COCO-CN. Specifically, we first extract an entity from the Chinese text and calculate the attention score of an image-entity pair. Here, we select the third layer of the text-image cross encoder following~\cite{ALBEF}.
\mmcr{Figure D in Appendix shows that \mmcr{R2D2} learns well to align text with the correct content inside the image.}



\begin{table}[t]
  \caption{\blue{Zero-shot results on image-text retrieval task.}}
  \centering
  \resizebox{0.48\textwidth}{!}{
  \begin{tabular}{c|lccccccc}
    \toprule
    &
    \multirow{2.5}{*}{Method} & 
    \multicolumn{3}{c}{Image-to-Text Retrieval} &
    \multicolumn{3}{c}{Text-to-Image Retrieval} &
    \multirow{2.5}{*}{R@M} 
    \\ 
    \cmidrule(r){3-5} \cmidrule(r){6-8}
    & & R@1 & R@5 & R@10 & R@1 & R@5 & R@10
    \\
    \midrule
    
    \multirow{4}{*}{\rotatebox{90}{\footnotesize Flickr30k-CN}}
    & $\text{CLIP}_\text{ViT-L}$~\cite{CLIP} & 75.0 & 94.5 & 97.7 & 51.8 & 78.6 & 85.9 & 80.6 \\
    & $\text{FILIP}_\text{ViT-L}$~\cite{filip} & \textbf{78.9} & 96.2 & 98.1 & 55.7 & 81.2 & 87.9 & 83.0 \\
    & $\text{Wukong}_\text{ViT-L}$~\cite{gu2022wukong} 
    &76.1 & 94.8 & 97.5 & 51.7 & 78.9 & 86.3 & 80.9 \\
    & $\text{\mmcr{R2D2}}_\text{ViT-L}$ 
    &77.6 & \textbf{96.7} & \textbf{98.9} & \textbf{60.9} & \textbf{86.8} &\textbf{92.7} & \textbf{85.6}  \\
    \midrule
    \multirow{4}{*}{\rotatebox{90}{COCO-CN}} 
    & $\text{CLIP}_\text{ViT-L}$~\cite{CLIP} & 51.0 & 80.0 & 89.7 & 48.7 & 76.8 & 86.4 & 72.1 \\
    & $\text{FILIP}_\text{ViT-L}$~\cite{filip} & 56.9 & 82.4 & 90.9 & 52.7 & 79.9 & 88.6 & 75.2 \\
    & $\text{Wukong}_\text{ViT-L}$~\cite{gu2022wukong} & 55.2 & 81.0 & 90.6 & 53.4 & 80.2 & 90.1 & 75.1  \\
    & $\text{\mmcr{R2D2}}_\text{ViT-L}$  & \textbf{63.3} & \textbf{89.3} & \textbf{95.7} & \textbf{56.4} & \textbf{85.0} & \textbf{93.1} &  \textbf{80.5}     \\
    \midrule

    \multirow{4}{*}{\rotatebox{90}{MUGE}}
    & $\text{CLIP}_\text{ViT-L}$~\cite{CLIP} &- &- &- & 43.3 & 69.2 & 78.4 & 63.6 \\
    & $\text{FILIP}_\text{ViT-L}$~\cite{filip} &- &- &- & 37.6 & 63.4 & 73.6 & 58.2 \\
    &  $\text{Wukong}_\text{ViT-L}$~\cite{gu2022wukong} 
    &- &- &- & 42.7 & 69.0 & 78.0 & 63.2  \\
    &  $\text{\mmcr{R2D2}}_\text{ViT-L}$ &- &- &- & \textbf{49.5} & \textbf{75.7} & \textbf{83.2} & \textbf{69.5}   \\
    \midrule
    \multirow{4}{*}{\rotatebox{90}{AIC-ICC}}
    & $\text{CLIP}_\text{ViT-L}$~\cite{CLIP} &16.8 &32.0 &39.8 &9.7 &21.1 &27.5 &24.5 \\
    & $\text{FILIP}_\text{ViT-L}$~\cite{filip} &20.6 &37.0 &45.4 &11.3 &24.3 &31.4 &28.3 \\
    & $\text{Wukong}_\text{ViT-L}$~\cite{gu2022wukong} & 18.2 & 34.5 & 42.4 & 8.8 & 20.3 & 27.3 & 25.3   \\
    & $\text{\mmcr{R2D2}}_\text{ViT-L}$  &\textbf{30.7} & \textbf{47.2} & \textbf{52.9} & \textbf{14.9} & \textbf{28.1} & \textbf{33.4} & \textbf{34.5}    \\
    \midrule
    \multirow{4}{*}{\rotatebox{90}{ICR}}
    & $\text{CLIP}_\text{ViT-L}$~\cite{CLIP} &30.3 &52.9 &61.6 &29.0 &51.9 &60.9 &47.8 \\
    & $\text{FILIP}_\text{ViT-L}$~\cite{filip} &27.3 &49.6 &58.3 &25.4 &48.5 &57.7 &44.5 \\
    & $\text{Wukong}_\text{ViT-L}$~\cite{gu2022wukong}& 35.1 & 58.2 & 66.3 & 33.7 & 58.0 & 66.5 & 53.0   \\
    & $\text{\mmcr{R2D2}}_\text{ViT-L}$   & \textbf{58.0} & \textbf{80.5} & \textbf{85.2} & \textbf{55.9} & \textbf{78.2} & \textbf{82.4} & \textbf{73.4}   \\
    \midrule

    \multirow{4}{*}{\rotatebox{90}{IQR}}
    & $\text{CLIP}_\text{ViT-L}$~\cite{CLIP} &24.3 &47.1 &56.2 &22.2 &45.2 &54.8 &41.6 \\
    & $\text{FILIP}_\text{ViT-L}$~\cite{filip} &21.9 &43.2 &52.8 &19.9 &42.0 &52.0 &38.6 \\
    & $\text{Wukong}_\text{ViT-L}$~\cite{gu2022wukong}& 26.1 & 48.9 & 58.1 & 24.9 & 48.1 & 57.7 & 44.0  \\
    & $\text{\mmcr{R2D2}}_\text{ViT-L}$  & \textbf{38.4} & \textbf{64.8} & \textbf{72.8} & \textbf{37.4} & \textbf{62.6} & \textbf{69.0} & \textbf{57.5}  \\
    \bottomrule
  \end{tabular}
  }
  \label{zero-shot}
\end{table}

\section{Conclusion}
\label{conclusion}
In this paper, we introduce a large-scale Chinese cross-modal benchmark called CCMB and a vision-language framework named \mmcr{R2D2}. CCMB includes a high-quality pre-training dataset \mmcr{Zero}, which is the largest Chinese cross-modal dataset, and five human-annotated downstream datasets, two of which are the largest Chinese vision-language downstream datasets and the first proposed datasets for the Chinese image-text matching task. \mmcr{R2D2 adopts a framework of pre-ranking + ranking for cross-modal learning, boosted with feature-guided distillation, target-guided distillation, and enhanced training.} 
After pre-training, \mmcr{R2D2} achieves state-of-the-art results on fine-tuning and zero-shot settings on twelve downstream datasets of five vision-language tasks.
We expect that the good cross-modal benchmark and framework will encourage a plethora of engineers to develop more effective methods in specific real-world scenarios.  

\mmcr{\textbf{Acknowledgement.} This work was supported by the National Key Research and Development Program of China (No.2018AAA010 0400).}

\bibliographystyle{ACM-Reference-Format}
\bibliography{egbib}

\newpage
\appendix

\setcounter{equation}{0}
\setcounter{table}{0}
\setcounter{figure}{0}



\newpage
\section{Details of Zero}
\label{details-pretrain}
\begin{figure*}[h]
    \renewcommand\thefigure{A}
    \centering
	\includegraphics[width=0.94\textwidth]{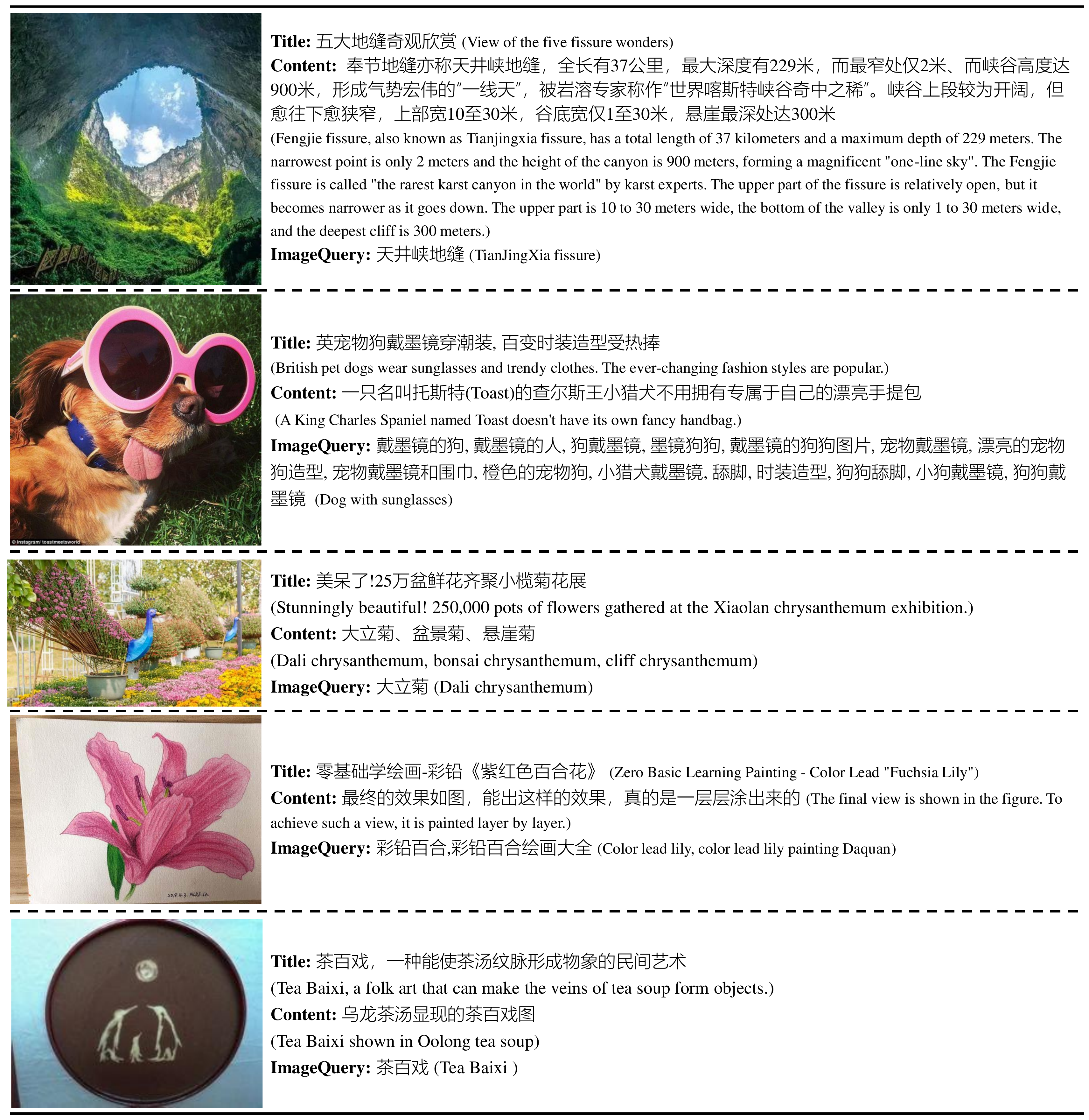}
	\caption{Examples of \mmcr{Zero}.} 
	\label{fig:pretraindata}
\end{figure*}

We illustrate several representative examples of \mmcr{Zero} in Figure \ref{fig:pretraindata}. 
There are 3 types of text fields associated with each image: ``Title'', ``Content'' and ``ImageQuery''. 


\section{Examples of the Proposed Downstream Datasets}
\label{details-downstream}
\begin{figure*}[h]
    \renewcommand\thefigure{B}
    \centering
	\includegraphics[width=0.94\textwidth]{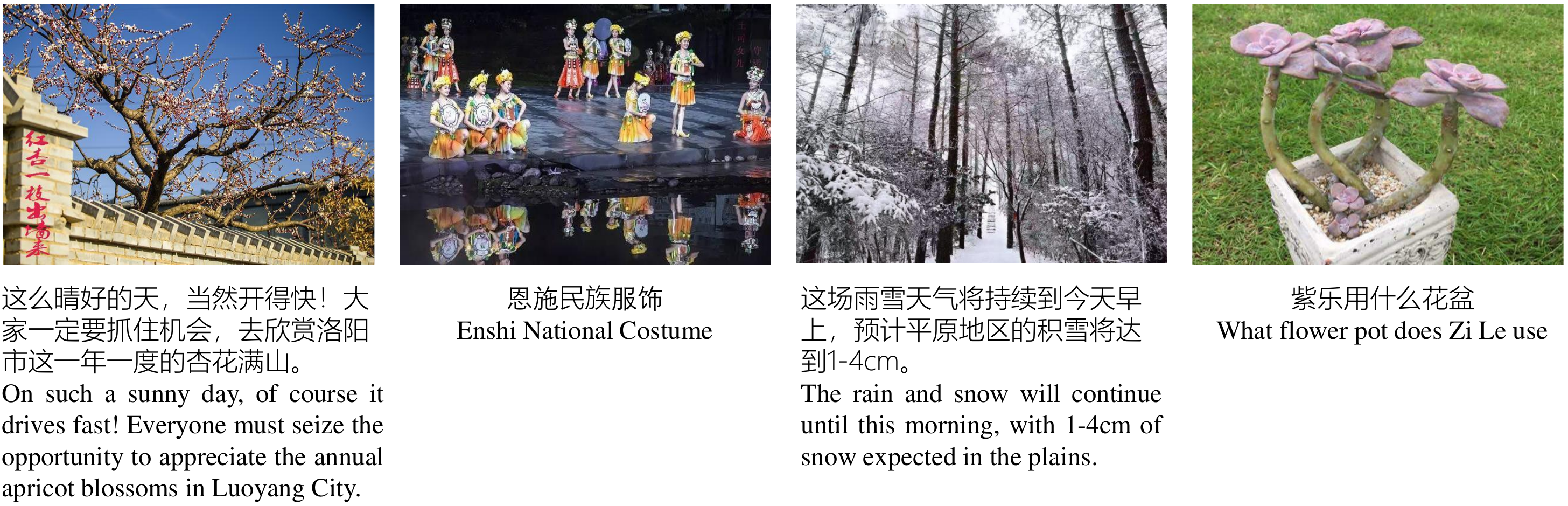}
    \caption{Image-text examples of ICM, IQM, ICR and IQR from left to right.}
	\label{fig:downstream} 
\end{figure*}

\begin{figure*}[h]
    \renewcommand\thefigure{C}
    \centering
	\includegraphics[width=0.94\textwidth]{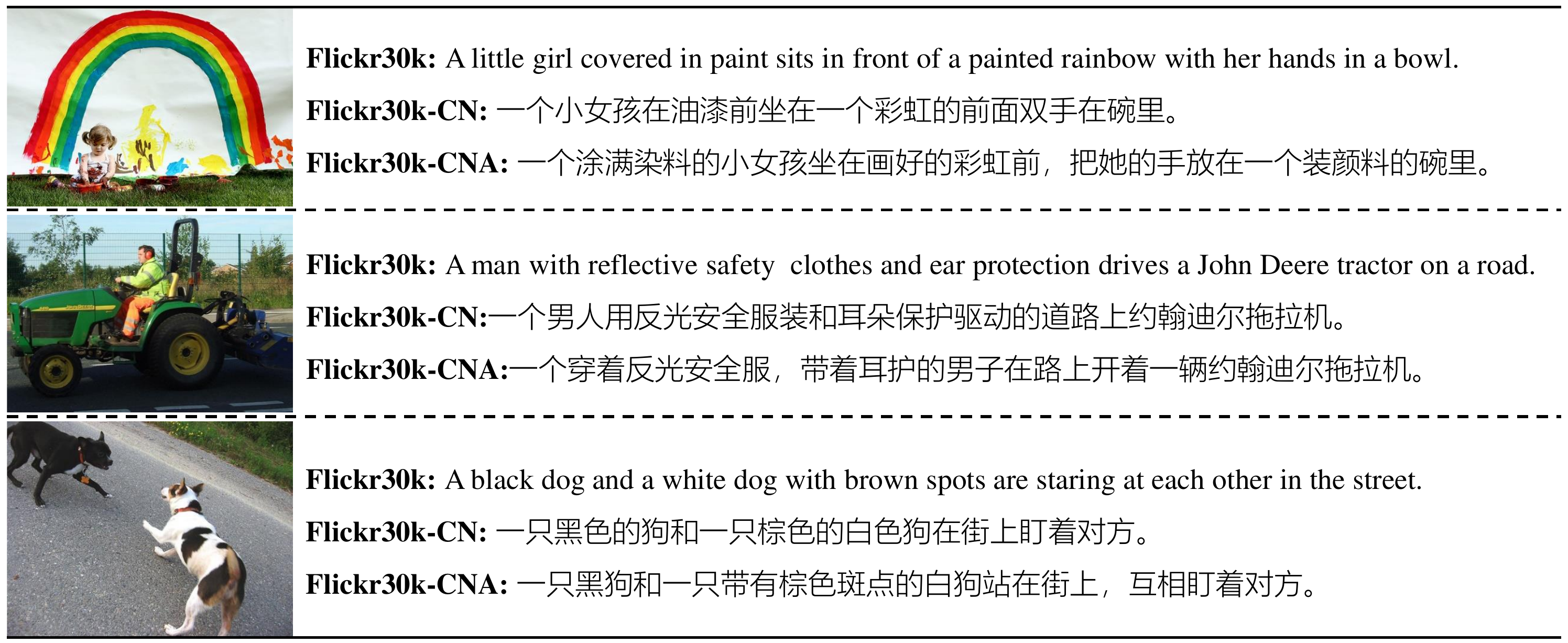}
	\caption{Comparisons of Flickr30k, Flickr30k-CN and our proposed Flickr30k-CNA.} 
	\label{fig:flickr}
\end{figure*}

\mmcr{Figure \ref{fig:downstream} illustrates examples of ICM, IQM, ICR, and IQR.}
Figure \ref{fig:flickr} highlights some cases of the difference between Flickr30k-CN and our proposed Flickr30k-CNA.

\section{More Implementation Details}
\label{details-finetune}
\textbf{Training process of target-guided distillation.}
\mmcr{
We use target-guided distillation following the settings of the momentum distillation in ALBEF~\cite{ALBEF}. Our goal is to generate soft targets to replace the ground-truth labels in Equation (\ref{ITCloss_gcpr}). During training, we perform the following processes.
i. For image and text, we use a momentum-updated encoder as the teacher model respectively, which contains the exponential-moving-average weights.
ii. We use the teacher models to obtain the corresponding image-text features and compute their similarity scores.
iii. We combine the above similarity scores with ground-truth labels via a coefficient parameter to generate the final soft targets.
iv. We replace the ground-truth labels in Equation (\ref{ITCloss_gcpr}) with the generated soft targets.
}

\textbf{Selection Strategy of Negative Pairs in Image-text Matching.}
\blue{We obtain hard negative samples by sampling in a mini-batch. Given an image in the mini-batch, we select the corresponding negative text by ranking the contrastive scores of the current batch. We choose the higher score except for the original positive text of the image. In this way, we construct one image-text negative pair for \mmljc{image-text matching} loss. The negative images of each text are similar to the description above.}

\textbf{Fine-tuning Strategy of Image-Text retrieval.}
We jointly optimize the GCPR loss (Equation \ref{ITCloss_gcpr}) and the \mmcr{FGR} loss (Equation \ref{FGRloss}). 
We extract the individual features of images and texts via our dual-stream encoder and
compute the similarity of all image-text pairs. 
\mmcr{During inference, we use the top-K strategy to rank the scores in the two cross encoders. We extract the individual features of images and texts via dual-stream encoders. For each image feature, we select the top-K candidate text features and construct K image-text pairs. We feed the K image-text pairs into two cross encoders to calculate similarity scores. We obtain two K-dimensional score matrices and average them to obtain a final K-dimensional score matrix for ranking.}
Here, we adjust the K on different downstream datasets.
We fine-tune the pre-trained model \jc{with 20 epochs} on 7 downstream datasets, including Flickr30k-CN, COCO-CN, AIC-ICC, MUGE, ICR, IQR, and Flickr30k-CNA.
K is set as 128, 256, 32, 64, 64, 64, 128, respectively. \jc{The batchsize is 32 and the learning rate is $1e^{-5}$.}

For both image-to-text retrieval (TR) and text-to-image retrieval (IR) tasks, we report Recall@1 (R@1), Recall@5 (R@5), Recall@10 (R@10), and Mean Recall (R@M).
For AIC-ICC and MUGE, we report their results on the validation sets, since their test sets are not released. For ICR and IQR, we also report the results on the validation sets in this paper.
For Flickr30k-CNA, we show the performance on the test set of Flickr30k-CN for a fair comparison in the main paper. 
For the remaining downstream datasets, we report the results on the test sets. Following~\cite{gu2022wukong}, we select the first 10,000 images with the corresponding 50,000 texts when testing on AIC-ICC. In particular, we only provide IR scores on MUGE since it only has IR settings.

\textbf{Fine-tuning Strategy of Image-Text Matching.}
This task predicts whether an image-text pair is matched or not. 
During fine-tuning, we only apply the \mmcr{FGR} loss (Equation \ref{FGRloss}). \jc{We fine-tune the models with 5 epochs using a batchsize of 64. The initial learning rate is $1e^{-5}$.}
Additionally, we report the results on the validation sets of ICM and IQM.

\textbf{Fine-tuning Strategy of Image Caption.}
\jc{Given an image, the goal of the image-caption task is to generate a caption to describe the image. Similar to Transformer\cite{Transformer}, the image-caption model consists of an encoder and a decoder, where the encoder aims to extract the embedding of the given image and the decoder generates tokens of the caption. In specific, we use the image encoder and the text-image cross encoder of \mmcr{R2D2} to initialize the image-caption encoder and decoder, respectively.
We fine-tune the image-caption model on the training split of AIC-ICC \cite{aic} with 20 epochs. The batchsize is 128 and the learning rate is $1e^{-4}$.
}

\textbf{Fine-tuning Strategy of Text-to-Image Generation.}
\jc{Text-to-image generation requires the model to generate an image corresponding to the input text.
Following DALL-E 2~\cite{DALLE2}, we build a generation model, including a CLIP-based module, a prior module and a decoder module. Specifically, the dual-stream weights of \mmcr{R2D2} are used to initialize the CLIP-based module. We fine-tune the CLIP-based module and fix it in the next step. Then, we train the prior module to generate image embeddings for given texts. Finally, we fix two former modules and train a diffusion decoder to invert the image embeddings to generate images. All three components of the generation model are fine-tuned on the ECommerce-T2I dataset with 20 epochs, respectively. The batchsize is 16 and the learning rate is $1e^{-4}$.}

\textbf{\blue{Fine-tuning Strategy of Zero-shot Image Classification.}}
\blue{Given an image, the zero-shot image classification task aims to predict the corresponding class label. Following \cite{gu2022wukong}, we use \mmcr{R2D2} to conduct zero-shot image classification task on ImageNet\cite{imagenet}. All the class labels in ImageNet are translated into Chinese.}

\section{Entity-conditioned Image Visualization}
\label{vis-appendix}
\begin{figure*}[h]
    \renewcommand\thefigure{D}
    \centering
	\includegraphics[width=0.8\textwidth]{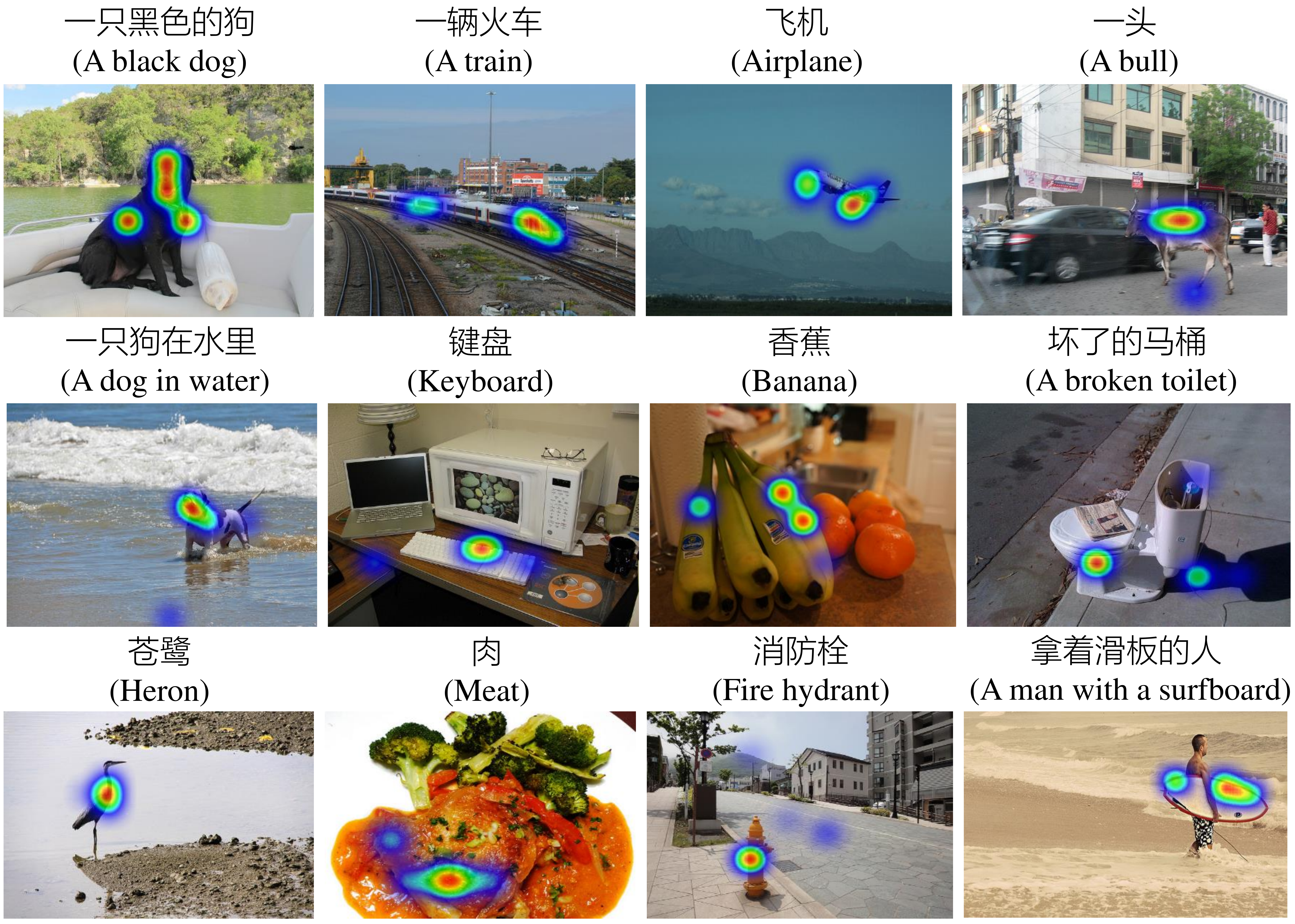}
	\caption{More Examples of entity-conditioned image visualization.} 
	\label{fig:visual_pos}
\end{figure*}

\begin{figure*}[h]
    \renewcommand\thefigure{E}
    \centering
	\includegraphics[width=0.8\textwidth]{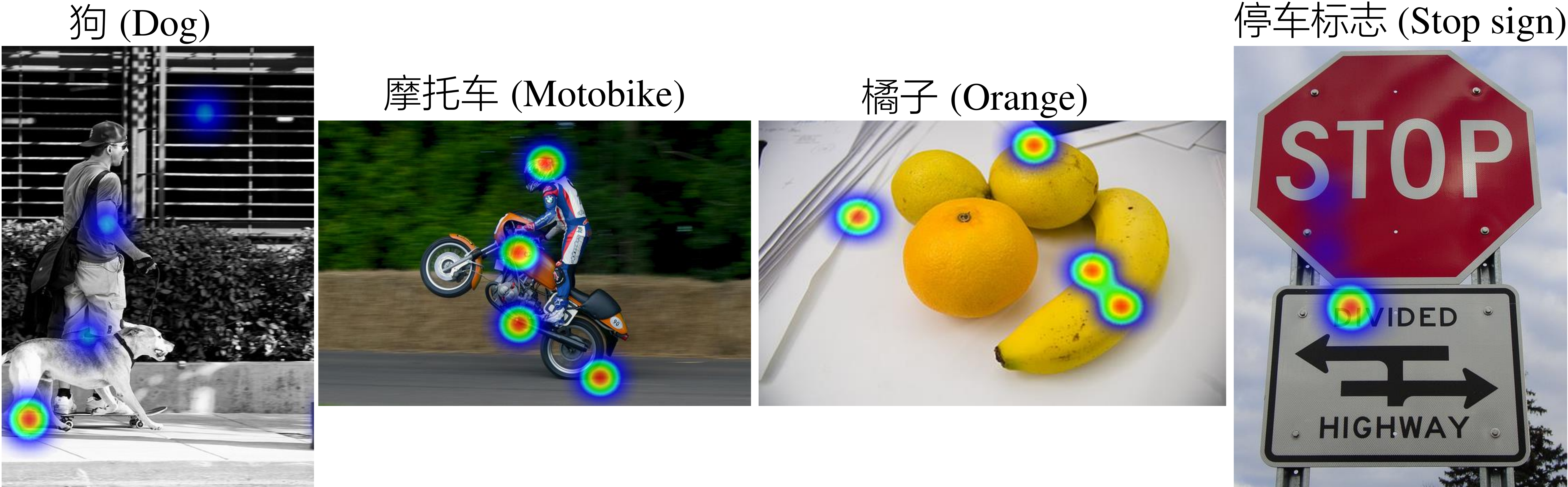}
	\caption{Bad cases of entity-conditioned image visualization.}
	\label{fig:visual_neg}
\end{figure*}

\mmcr{In this experiment, we visualize the attention map of images on COCO-CN.}
From Figure \ref{fig:visual_pos}, 
\mmcr{R2D2} has the ability to capture the salient areas when given an image with complex backgrounds,
such as the images of ``A train'' and ``A bull''.
Moreover, we analyze some bad cases in Figure \ref{fig:visual_neg}. We find that the attention score is disturbed when two adjacent entities are present in an image. This phenomenon is particularly evident for objects with similar colors or categories.


\end{document}